\newcommand{\CLASSINPUTtoptextmargin}{2.05cm}
\newcommand{\CLASSINPUTbottomtextmargin}{4.4cm}
 \documentclass[conference, a4paper]{IEEEtran}
\IEEEoverridecommandlockouts
\usepackage[numbers]{natbib}
\usepackage{amsmath,amssymb,amsfonts}
\usepackage{algorithmic}
\usepackage{graphicx}
\usepackage{textcomp}
\usepackage{subcaption, caption}
\usepackage{url}
\usepackage{multicol}
\usepackage{multirow}
\usepackage{xcolor}
\usepackage{enumitem}
\usepackage{placeins}
\usepackage{fancyhdr}
\newcommand{\EXCLUDE}[1]{}

\def\BibTeX{{\rm B\kern-.05em{\sc i\kern-.025em b}\kern-.08em T\kern-.1667em\lower.7ex\hbox{E}\kern-.125emX}}

\fancypagestyle{firstpage}{%
 \fancyhf{}%
 \fancyhead[L]{Conference on Digital Platforms and Societal Harms 2024 (DPSH 2024)\\
  \textit{October 14-15, 2024, Washington, DC, USA}}

  \fancyfoot[L]{%
  \normalsize{979-8-3503-8359-1/24/\$31.00 ~\copyright2024 IEEE}
 }%
}

\begin{document}

 \title{Modeling offensive content detection for {TikTok} \\
{
}
}

\author{\IEEEauthorblockN{1\textsuperscript{st} Kasper Cools}
\IEEEauthorblockA{
\textit{Vrije Universiteit Brussel,} \\
\textit{Belgian Royal Military Academy} \\
kasper.cools@vub.be}

\and
\IEEEauthorblockN{2\textsuperscript{nd} Gideon Mailette de Buy Wenniger}
\IEEEauthorblockA{
\textit{Open University of the Netherlands,}\\ 
\textit{University of Groningen}\\
gideon.maillette.de.buy.wenniger@ou.nl}
\and
\IEEEauthorblockN{3\textsuperscript{rd} Clara Maathuis}
\IEEEauthorblockA{
\textit{Open University of the Netherlands}\\ 
clara.maathuis@ou.nl}
}

\maketitle
\thispagestyle{firstpage}

\begin{abstract}
   The advent of social media transformed interpersonal communication and information consumption processes. This digital landscape accommodates user intentions, also resulting in an increase of offensive language and harmful behavior. Concurrently, social media platforms collect vast datasets comprising user-generated content and behavioral information. These datasets are instrumental for platforms deploying machine learning and data-driven strategies, facilitating customer insights and countermeasures against social manipulation mechanisms like disinformation and offensive content. Nevertheless, the availability of such datasets, along with the application of various machine learning techniques, to researchers and practitioners, for specific social media platforms regarding particular events, is limited. In particular for TikTok, which offers unique tools for personalized content creation and sharing, the existing body of knowledge would benefit from having diverse comprehensive datasets and associated data analytics solutions on offensive content. While efforts from social media platforms, research, and practitioner communities are seen on this behalf, such content continues to proliferate. This translates to an essential need to make datasets publicly available and build corresponding intelligent solutions. On this behalf, this research undertakes the collection and analysis of TikTok data containing offensive content, building a series of machine learning and deep learning models for offensive content detection. This is done aiming at answering the following research question: "How to develop a series of computational models to detect offensive content on TikTok?". To this end, a Data Science methodological approach is considered, 120.423 TikTok comments are collected, and  on a balanced, binary classification approach, F1 score performance results of 0.863 is obtained.
\end{abstract}

\section{Introduction}

The ubiquity of various social media platforms has brought a wide spectrum of implications and consequences for both individuals and society as a whole. On the one hand, these platforms have redefined the landscape of interpersonal communication, a fact that established new ways of communication and interaction across geographic boundaries. sAt the same time, social media platforms serve as dynamic forums for community building, networking, and catalyzing social movements~\cite{tiktokyoung}. On the other hand, despite these benefits, challenges like the dissemination of mis/disinformation mechanisms, the perpetuation of cyber bullying, and security and privacy breaches, reveal the negative side of social media engagement. Among these platforms, TikTok rapidly emerged as a dynamic and influential social media platform, captivating over a billion users worldwide~\cite{tiktokstatistics} with its innovative short-form video format, user-friendly interface, and integration of music and visual effects~\cite{feldkamp21, 2021PsychologyTikTok}. TikTok is a platform for community formation, encourages individuals to express themselves, and fosters collective engagement around common interests and trends through the so-called “echo chambers” effect~\cite{CiaranOConnor2021,cinelli2021echo}. Nevertheless, through its widespread use, the formation and proliferation of offensive and harmful content is also seen on this platform given the use of fast engagement-driven algorithms and limited content moderation~\cite{tiktokmoderation}.    

As a significant portion of TikTok users comprises young individuals like Gen-Z (29.5\%) and aged between 10 and 19 (32.5\%), the proliferation of offensive content holds heightened significance due to its potential impact on users’ perceptions, behaviors, and overall well-being~\cite{CiaranOConnor2021}. In this sense, a tendency to shield users from opinion-challenging information to encourage them to adopt more extreme views is seen~\cite{kitchens2020understanding}. Such content can also have a substantial impact on the mental health of young individuals~\cite{bucur2021exploratory}. Consequently, addressing the surge in offensive content on TikTok became imperative to upholding societal and community standards and safeguarding its users' safety, security, and privacy. To tackle this threat, a collective joint effort is required, nonetheless, this process is in its infancy. Accordingly, governance efforts include the establishment of regulatory frameworks and industry standards, practitioner efforts comprise the development of platform mechanisms and active engagement of content moderators, and researcher efforts represent a pivotal role in advancing understanding of the underlying mechanisms driving the proliferation of offensive content and the development of innovative approaches and technologies for offensive content detection. While various mechanisms and solutions are proposed for dealing with social media threats like disinformation and cyber bullying, seeing the fast pace of spread and increase in complexity of offensive content, additional technical efforts are necessary. To support these initiatives, the release of open publicly available datasets and the development of various effective AI-based solutions play an important role as this would facilitate the creation of more robust and innovative offensive content detection models and mechanisms tailored to this platform. 

To contribute to existing efforts tackling this threat, this research aims at building a TikTok dataset~\footnote{For access to the dataset used in this study, interested researchers can contact one of the authors via email.} and a series of deep learning and machine learning models for detecting offensive content, leveraging advanced Natural Language Processing (NLP) techniques that allow understanding the context and intention behind users’ content. Hence, the following research question is formulated: “How to develop a series of computational models for detecting offensive content on TikTok?”. To answer this question, the Data Science research methodology~\cite{datasciencemethodology} is applied by merging literature review, field expertise, and development of deep learning and machine learning models. To this end, the following contributions of this research are considered: 

\begin{itemize}
	\item A TikTok dataset based on 120.423 comments, contains data between April to July 2022, and is collected using a combination of web scraping techniques.
	\item Data insights obtained through comprehensive textual processing and analysis using machine learning and NLP techniques like trigrams, TF-IDF, word clouds, and topic modelling. This approach allows a quantitative investigation of linguistic patterns and emojis associated with offensive language and assessing the prevalence of specific words, word combinations, and emojis in TikTok offensive content.
    \item A set of deep learning and machine learning models built for offensive content detection using the BERT (Bidirectional Encoder Representations from Transformers), logistic regression (LR), and na{\"\i}ve bayes (NB) algorithms which have been previously successfully used in similar tasks. Out of these, the BERT models are the best-performing ones, with F1 scores between 0.851 and 0.863.
\end{itemize}

The remainder of this article is structured as follows. Section~\ref{section:related-work} discusses relevant studies to this research. Section~\ref{section:data-collection-annotation} discusses the data collection process. Section~\ref{section:data-analysis} explores the dataset compiled and proposed in this study. Section~\ref{section:design} presents the pre-processing and modelling processes and Section~\ref{section:results} discusses the results obtained. Finally, Section~\ref{section:conclusion} presents concluding remarks and future research perspectives.



\section{Related work}\label{section:related-work}

Research in the field of harmful content classification has seen significant advancements with methodologies employing both machine learning and deep learning techniques.
Pradhan et al.~\cite{pradhan2020review} highlighted the limitations of methods that heavily rely on dictionaries, pointing out the need for more robust techniques. On the other hand, Davidson et al.~\cite{davidson2017automated} proposed a multi-class classification model capable of distinguishing hate speech, demonstrating the potential of machine learning in this domain.
A study by Alatawi et al.~\cite{alatawi2021detecting} concentrated on identifying white supremacist hate speech using a BiLSTM model, showcasing the effectiveness of deep learning techniques in specific hate speech detection.
Additionally, the fine-tuning of existing models, such as BERT, has also been explored extensively. Caselli et al.~\cite{caselli2020hatebert} introduced HateBERT, a fine-tuned version of BERT specifically designed for hate speech detection. Similarly, Apoorva et al.~\cite{ApoorvaK2022} developed a series of models for detecting cyber bullying content, while Darmawan et al.~\cite{Darmawan2023} devised a multi-label hate speech detection model for the Indonesian language utilizing indoBERT.
Furthermore, Myilvahanan et al.~\cite{Myilvahanan2023} integrated machine learning with BERT for sarcasm detection, and Prameswari et al.~\cite{Prameswari2023} developed a model for identifying cyber bullying on TikTok using BERT. These studies highlight the versatility and effectiveness of BERT in various aspects of harmful content detection.
Duong et al.~\cite{Duong2022} presented HateNet, a model that employs a Graph Convolutional Network classifier and a weighted DropEdge-based stochastic regularization technique for hate speech detection.
Similarly, Hernandez et al.~\cite{hernandez2021bert} developed a BERT-based model for detecting hate speech in short-form TikTok video content. To achieve this, they collected and manually transcribed 1000 TikTok videos in the Filipino language. Singh et al.~\cite{Singh2023} fine-tuned a custom BERT implementation named RoBERTA for hate speech detection based on the Hate Speech and Offensive Content Identification (HASOC) dataset. Research by Samee et al.~\cite{Samee2023} further confirmed that BERT-based models can outperform other deep learning solutions. This research proposes cyber bullying detection models using data from Twitter, Wikipedia Talk pages, and YouTube. Adding to the body of multilingual offensive content detection, Ranasinghe and Zampieri~\cite{ranasinghe2020multilingual} introduced a classifier model trained and fine-tuned on an XML-R transformer architecture. They used the Offensive Language Identification Dataset (OLID), which contains over 14,000 manually annotated Tweets following a three-level taxonomy. Along these lines, Turjya et al.~\cite{turjya2023multilingual} trained and fine-tuned a BERT-based multilingual hate speech and offensive language detection model, further emphasizing the importance of multilingual approaches in this research area.
Nevertheless, more efforts need to be dedicated to building intelligent solutions for harmful and offensive content detection on TikTok. 

\section{Data collection and annotation}\label{section:data-collection-annotation}
To account for TikTok's unique features and social context, a custom dataset was compiled. During this research, TikTok made efforts to enhance transparency and provide researchers access to publicly available data through their Research API\footnote{\url{https://developers.tiktok.com/products/research-api}}.
Before this, however, there was no straightforward method to automatically retrieve information in this platform. Previously, data collection from TikTok was limited to web scraping. Nonetheless, solutions like TikAPI and the unofficial Python TikTok API have automated this process. For this research, BrightData’s automated data scraping service was used.
The TikTok algorithm, which personalizes content feeds based on user interactions, can result in varied data across users and sessions~\cite{klug2021trick}. Therefore, data from multiple users and sessions were aggregated in this research.

In an effort to guide the search algorithm towards more offensive content, a lexicon known as WeaponizedWord (WW) was utilized. This lexicon is specifically designed to identify and categorize hate speech, and includes terms that fall under four categories: discriminatory, derogatory, threatening, and watchwords.
Discriminatory terms such as ‘mong’ and ‘retard’, which are included in the lexicon, served as examples of words used for searching. The use of such terms is strictly for research purposes and to illustrate the type of language that the lexicon can identify.
The WW lexicon was used to guide the search algorithm towards relevant content. However, the algorithm’s complexity and unpredictability limited the effectiveness of this approach.

Analyzing offensive messages involves examining both user posts and their replies, as offensive language often appears in online comment threads. Comments were collected from all the retrieved posts using a Python script built with Selenium Webdriver, resulting in a total of $120,423$ comments. The post structure and relationships were preserved allowing for the reconstruction of conversational tree structures for future research.

Instead of using machine learning models, which can be biased~\cite{mehrabi2021survey}, messages were manually examined and labeled based on criteria from prior research by Zampieri et al.~\cite{zampieri2020semeval} and included:
\begin{itemize}
    \item Insults or threats targeted towards an individual or a group;
    \item Inappropriate language, insults, or threats;
    \item Explicit or implicit targeting of people based on ethnicity, gender, sexual orientation, religious belief, or other common characteristics.
\end{itemize}

Using specific criteria, the data was binary labeled as either ‘Offensive’ or ‘Not Offensive’. During the textual analysis, accurately understanding and identifying offensive messages was challenging. This was primarily due to frequent misspellings and disregard for punctuation often observed in messages on online social media platforms. These linguistic variations could be attributed to factors like the writer’s age, geographical upbringing ~\cite{hasyim2019linguistic}, and character limitations imposed by the platform (e.g., TikTok restricts posts to 150 characters)~\cite{chen2012detecting, gouws2011contextual}. A total of $120,423$ comments were collected. Out of these, $2,034$ unique comments were identified as offensive, while $75,650$ comments were deemed non-offensive. To create a balanced dataset, an equal number of non-offensive comments were randomly selected. This resulted in a dataset of $4,068$ comments, which was used for data analysis and model building.


\section{Data analysis}\label{section:data-analysis}
For performing analyses on the dataset, different pre-processing steps need to be carried out depending on the type of analysis. The following sections describe in more detail the different pre-processing steps taken. As every analysis aims to measure a different aspect or feature of the dataset, not all steps were necessarily applied to each analysis.

\subsection{Stop words, Lemmatization, and Punctuation}
Stop words \cite{9074166,sarica2021stopwords}, which do not add additional meaning to text, were removed using the
Natural Language Toolkit (NLTK) module \cite{sarica2021stopwords}.
The default stop words removal list was extended with the following  shorthand words, specifically featured in the dataset:  ``u'', ``ur'', ``cause'', ``gonna'', ``im'', ``gon'', ``cant''.

Lemmatization converts the words to their meaningful base form (lemma). Lemmatization was chosen over stemming (which allows to reduce words with the same stem to their common form), as it achieves better accuracy~\cite{balakrishnan2014stemming}.
This was performed using the NLTK WordNetLemmatizer\footnote{\url{https://nltk.org/_modules/nltk/stem/wordnet.html}}.


Removing special characters and punctuation markers can be important for certain analyses, as these might not add extra meaning or value. 
However, since social media users often use (textual) emojis to convey emotions, removing them may affect the output of the analysis~\cite{8282676,katz2022gen}. Therefore, this research uses a set of regular expressions to remove the punctuation and special characters.

\subsubsection{Word clouds}\label{section:wordclouds}

\begin{figure}[tb]
     \centering
     \begin{subfigure}[b]{0.48\textwidth}
         \centering
        \includegraphics[scale=0.20]{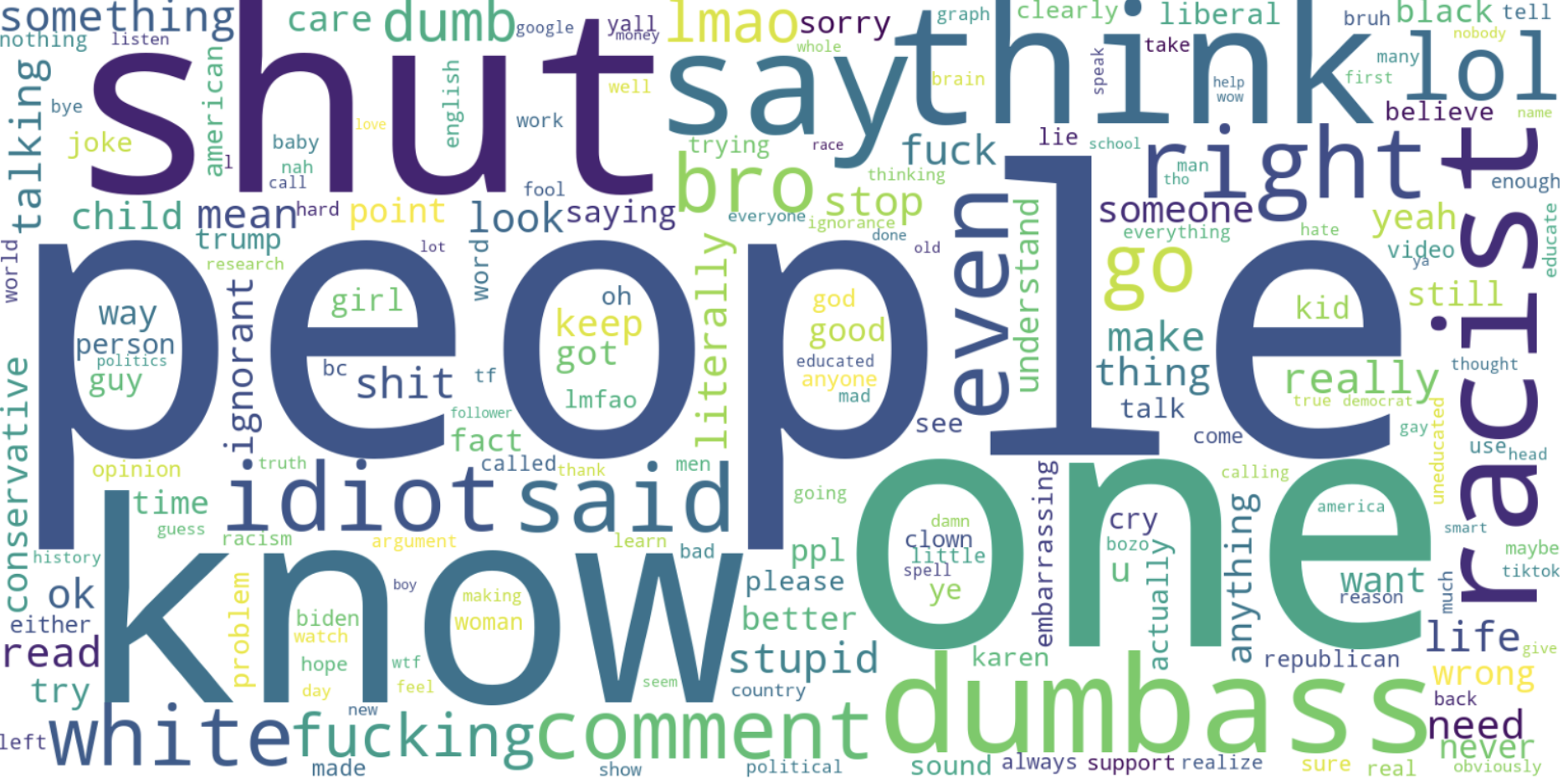}
		\caption{based on individual words labeled offensive}\label{fig:wordcloud}
     \end{subfigure}
     \hfill
     \begin{subfigure}[b]{0.48\textwidth}
         \centering
         \includegraphics[scale=0.20]{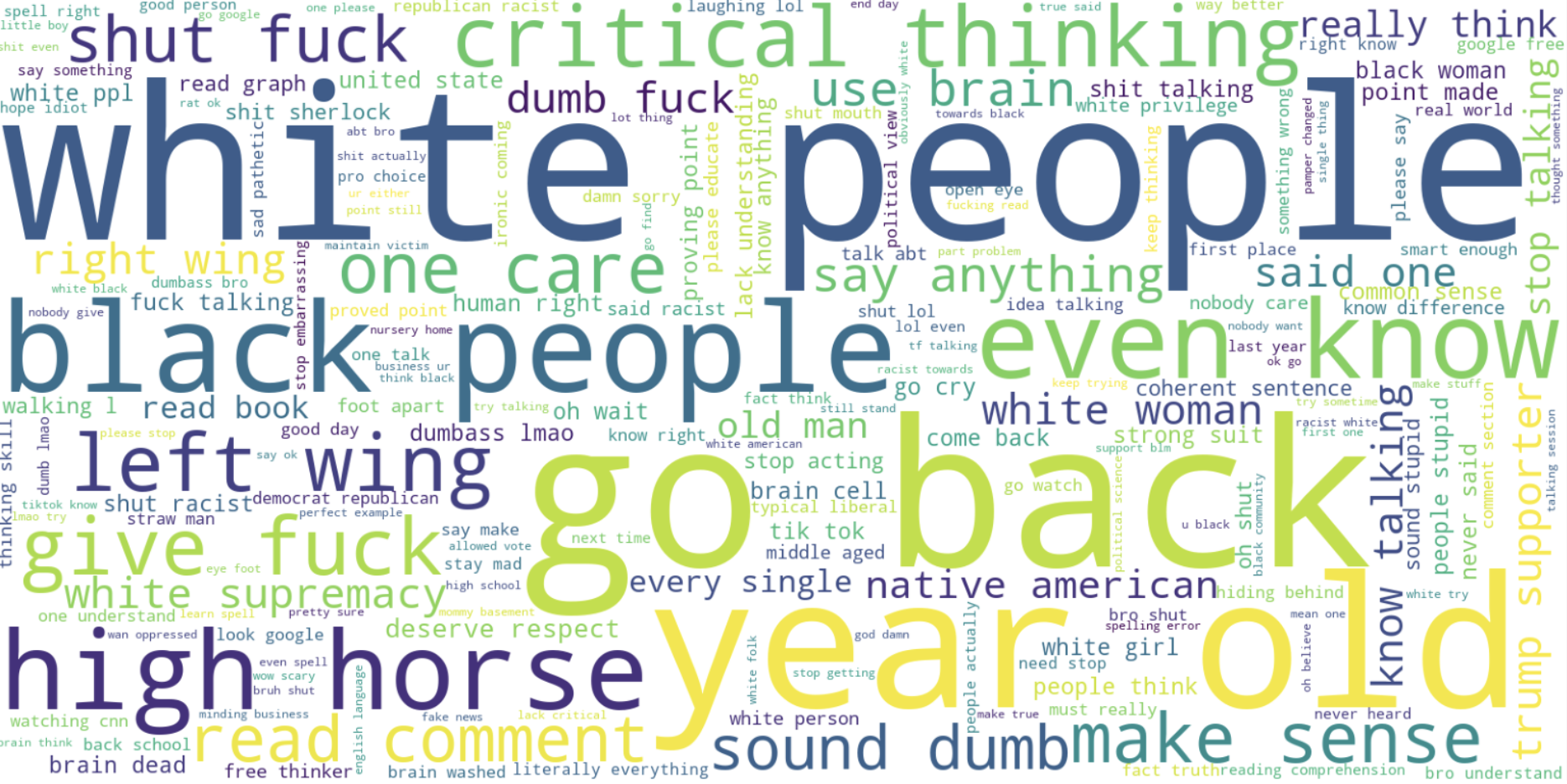}
		  \caption{bi-grams based on the dataset label as offensive}\label{fig:wordcloud-bigrams}
     \end{subfigure}
        \caption{Word clouds}
        \label{fig:word-clouds}
\end{figure}

Word clouds 
facilitate the visualization of the most frequently words used by individuals when using offensive language~\cite{HeimerlFlorian2014}. 
To this end, Figure~\ref{fig:wordcloud} provides an overview of individual words which are most used in the corpus of operandi; showing that the words ``People'', ``One'', ``Know'' bear the greatest weight in the chart. This chart indicates that most of the offensive comments use the word ``people'', which could refer to intercultural or political discussions. Other prominent words are insults such as ``racist'', ``dumbass'', and words like ``shut''. 


The following examples illustrate the actual use of these individual words in the dataset:

\begin{itemize}[itemsep=0.5ex]
	\item ``shut up karen i'll kneel to the american flag burning periodt''
	\item ``They only teach you what they want you to know idiot thats why so many politicians are corrupt you gave yourself lies to fill empty space in your head''
\end{itemize}

A bi-gram word  cloud, depicted in Figure~\ref{fig:wordcloud-bigrams} identifies the most commonly used word pairs. Substantial are the pairs ``white people'' and ``black people'', which could indicate that many discussions are about ethnicity or race. Furthermore, the middle left of the cloud shows the words ``left wing'', which indicates that the dataset may contain quite a few comments about political opinions. Other combinations like ``year old'' and ``critical thinking'' are commonly used as a way to direct attention to someone's level of intelligence. 
\begin{itemize}[itemsep=0.5ex]
	\item ``Are you 14? You reason like a 14 year old girl''
	\item ``all the right wing ppl lack extreme critical thinking''
\end{itemize}
The word pair ``high horse'' on the other hand may refer to those who, according to the writer, think they are better or know better than others when it comes to a certain topic (e.g. ``so get off your racist high horse and do better'').
\begin{itemize}[itemsep=0.5ex]
	\item ``It was a mistake Karen maybe go back to school to get your IQ back??''
	\item ``I bet you don't even know what the BLM movement stands for. Dumb lib. It goes farther then striving for equality for black people.''
\end{itemize}

Generally, most of the pairs are easy to understand and relate to a particular context. Some, however, might not be that obvious, such as ``even know'' or ``go back''. These are not merely insults but combinations that need more context:
\begin{itemize}[itemsep=0.5ex]
	\item ``We've always had a foreign dependence on oil, and half you idiots don't even know what country imports the most oil to us''
	\item ``Maybe you need to go back to school and learn some writing skills before typing comments cannot even get a message out correctly.''
\end{itemize}

\begin{figure}[t]
 \begin{subfigure}[b]{0.48\textwidth}
       \centering
		\includegraphics[scale=0.25]{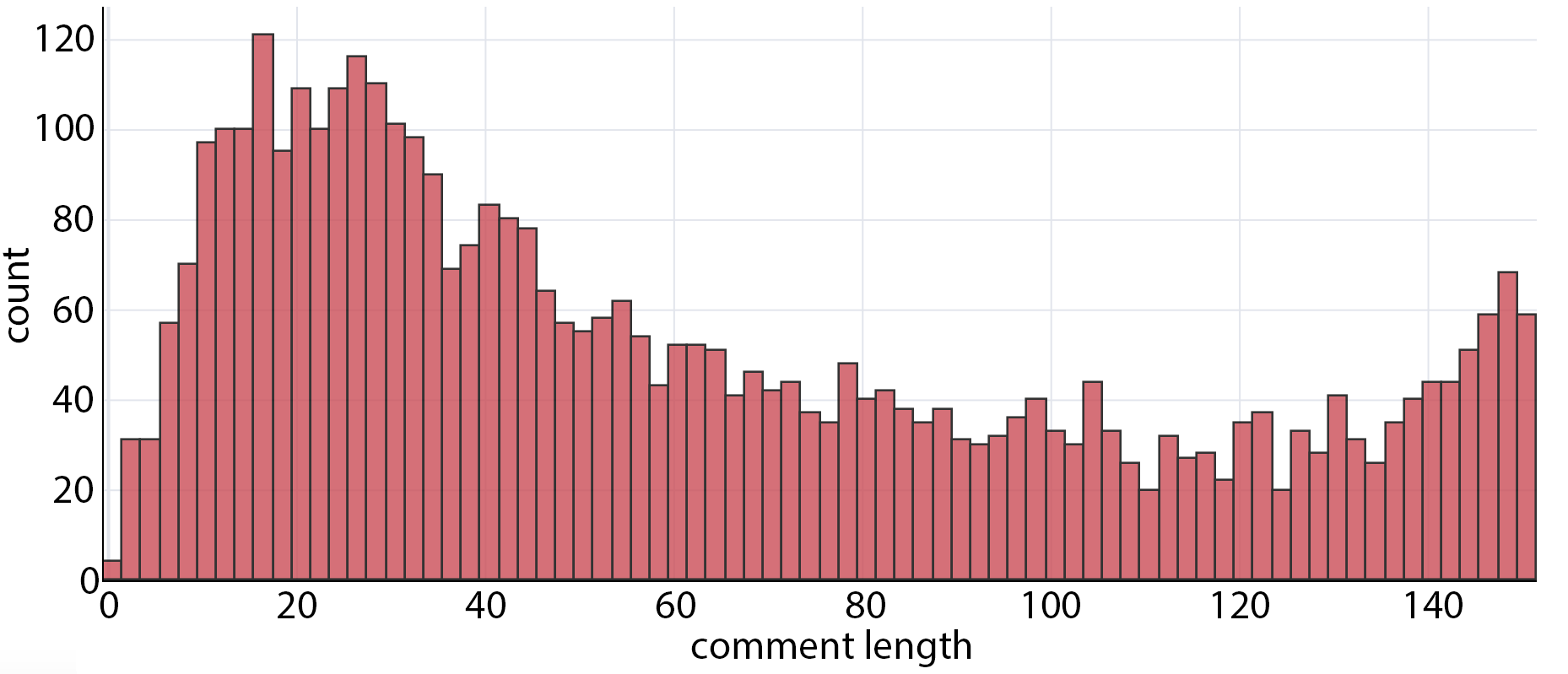}
		\caption{Offensive and non-offensive comments}
		\label{fig:general-comment-length-distribution}
 \end{subfigure}
 \hfill
  \begin{subfigure}[b]{0.48\textwidth}
          \centering
		\includegraphics[scale=0.25]{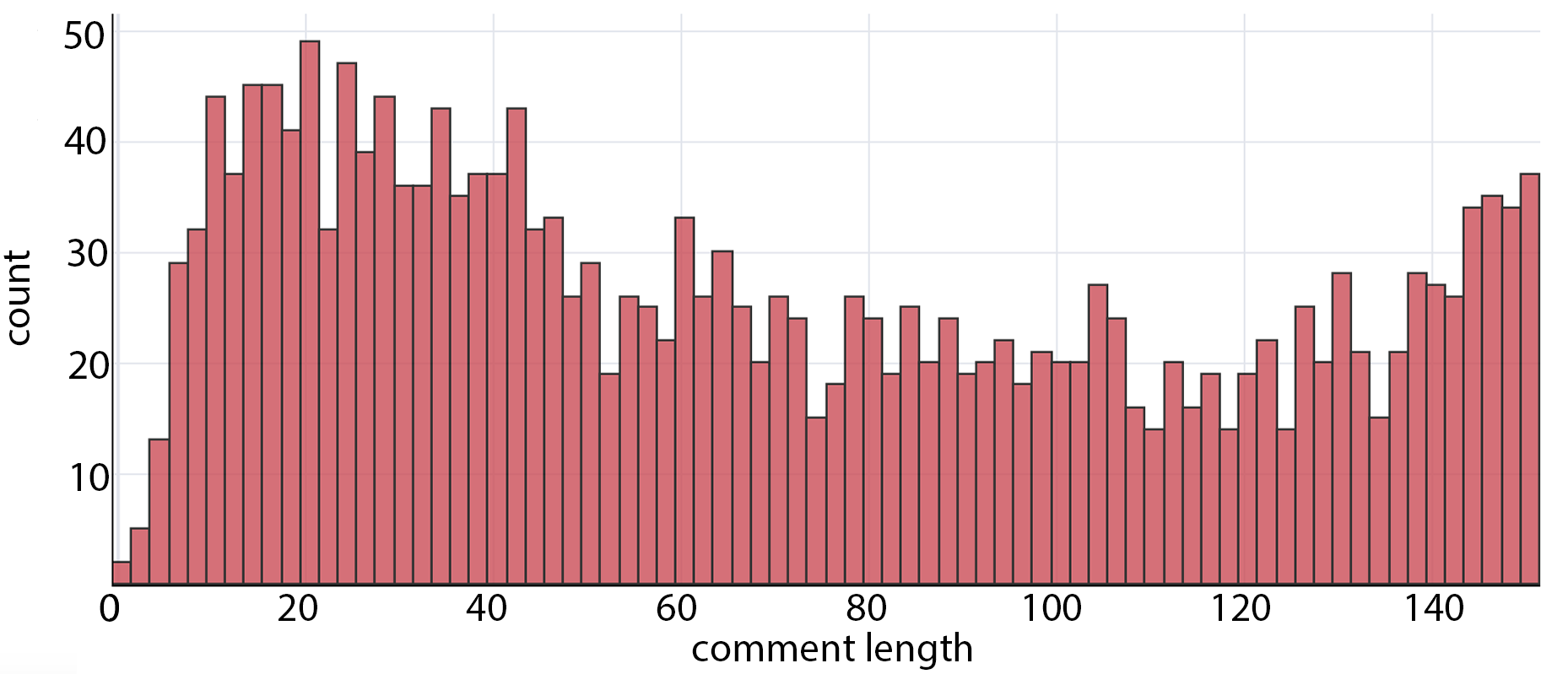}
		\caption{Offensive comments}
		\label{fig:offensive-comment-length-distribution}
 \end{subfigure}
\caption{Comment length distributions}
		\label{fig:comment-length-distribution}
\end{figure}

\begin{figure*}[t]
     \centering
     \begin{subfigure}[b]{0.40\textwidth}
         \centering
        \includegraphics[scale=0.25]{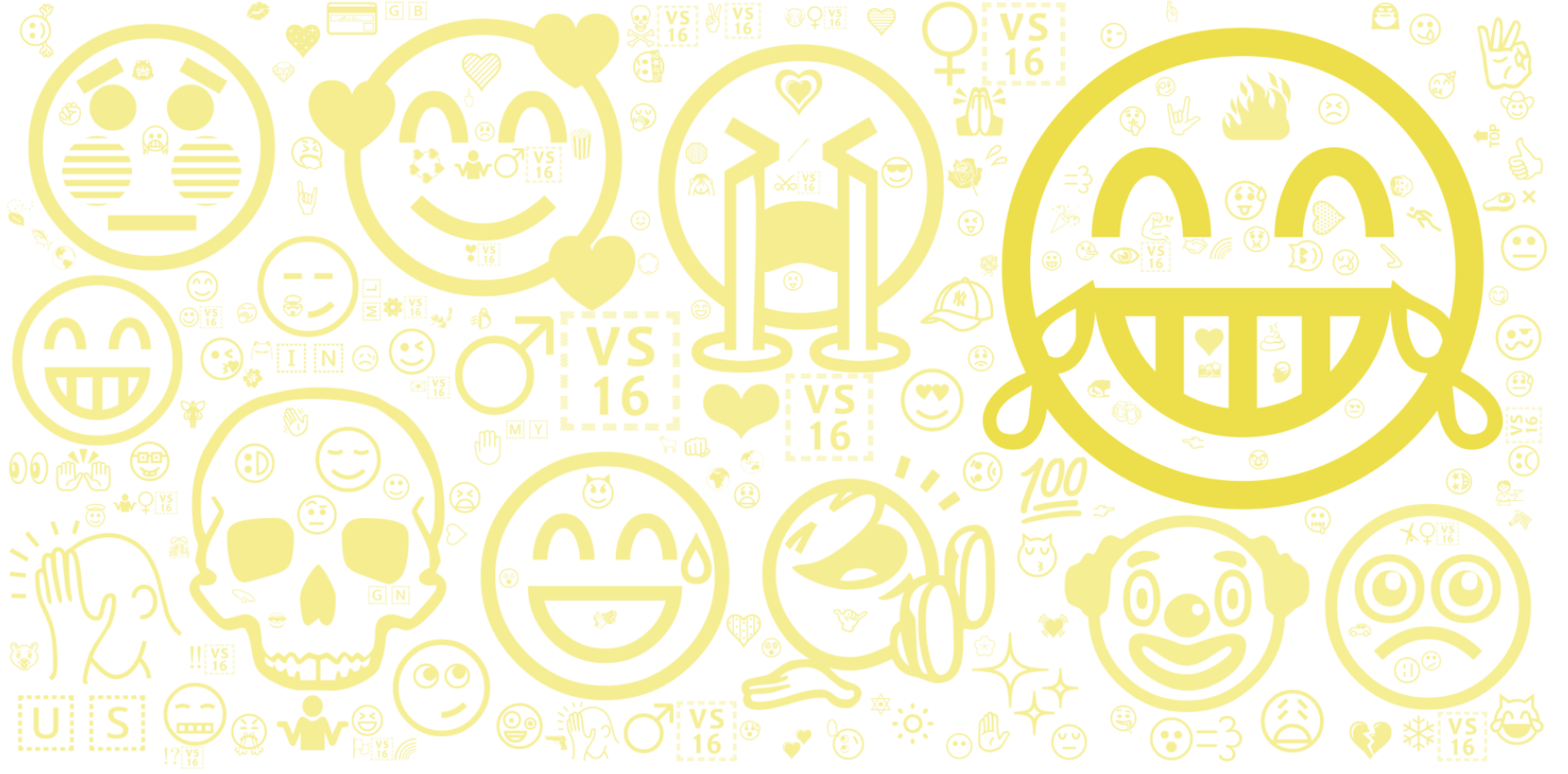}
		\caption{offensive and non-offensive comments}
		\label{fig:most-freq-used-emojis-cloud}
     \end{subfigure}
     \begin{subfigure}[b]{0.48\textwidth}
         \centering
         \includegraphics[scale=0.25]{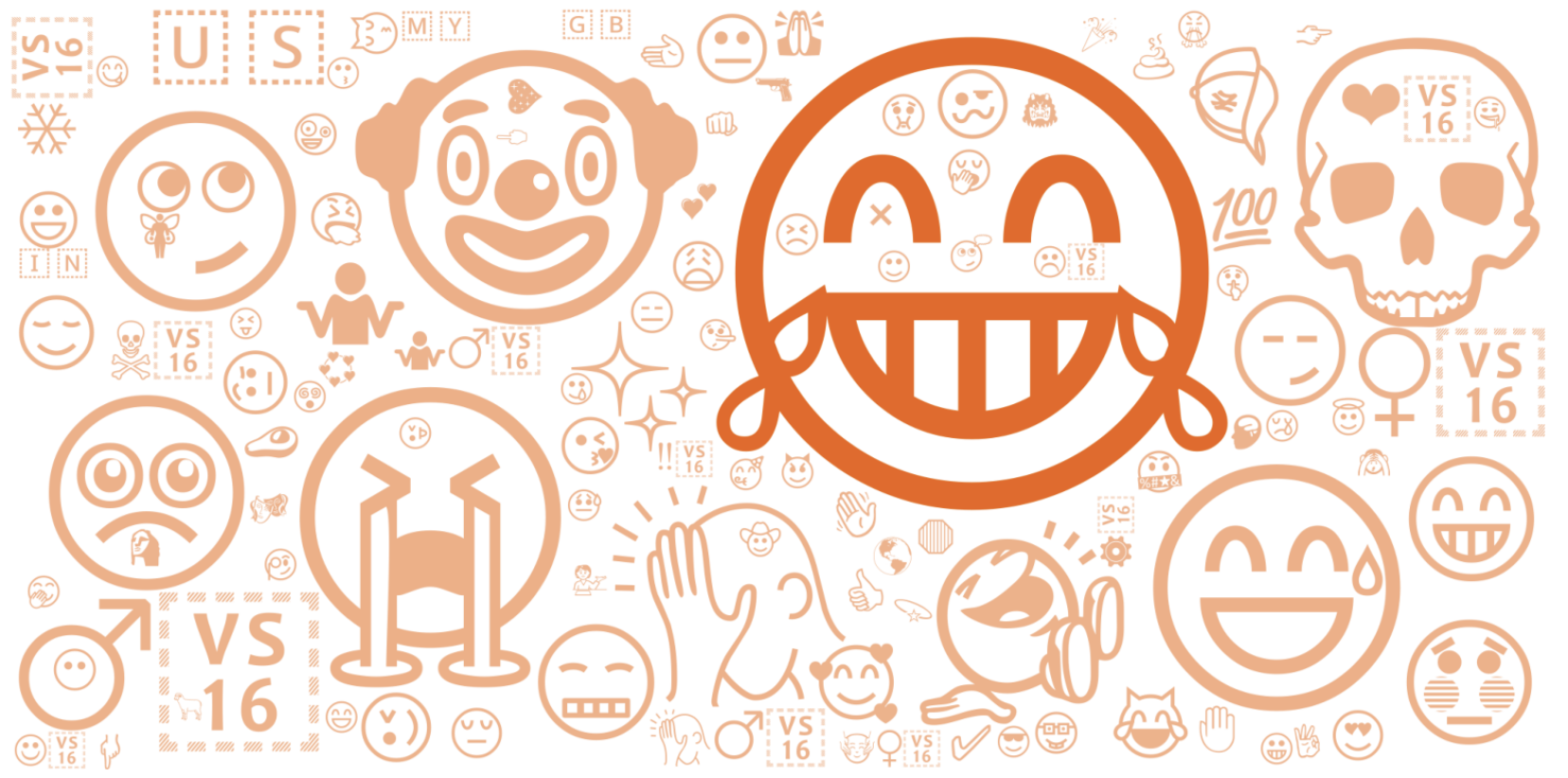}
		\caption{offensive comments}
		\label{fig:most-freq-used-offensive-emojis-cloud}
     \end{subfigure}
        \begin{subfigure}[b]{0.48\textwidth}
        \centering
        \includegraphics[scale=0.24]{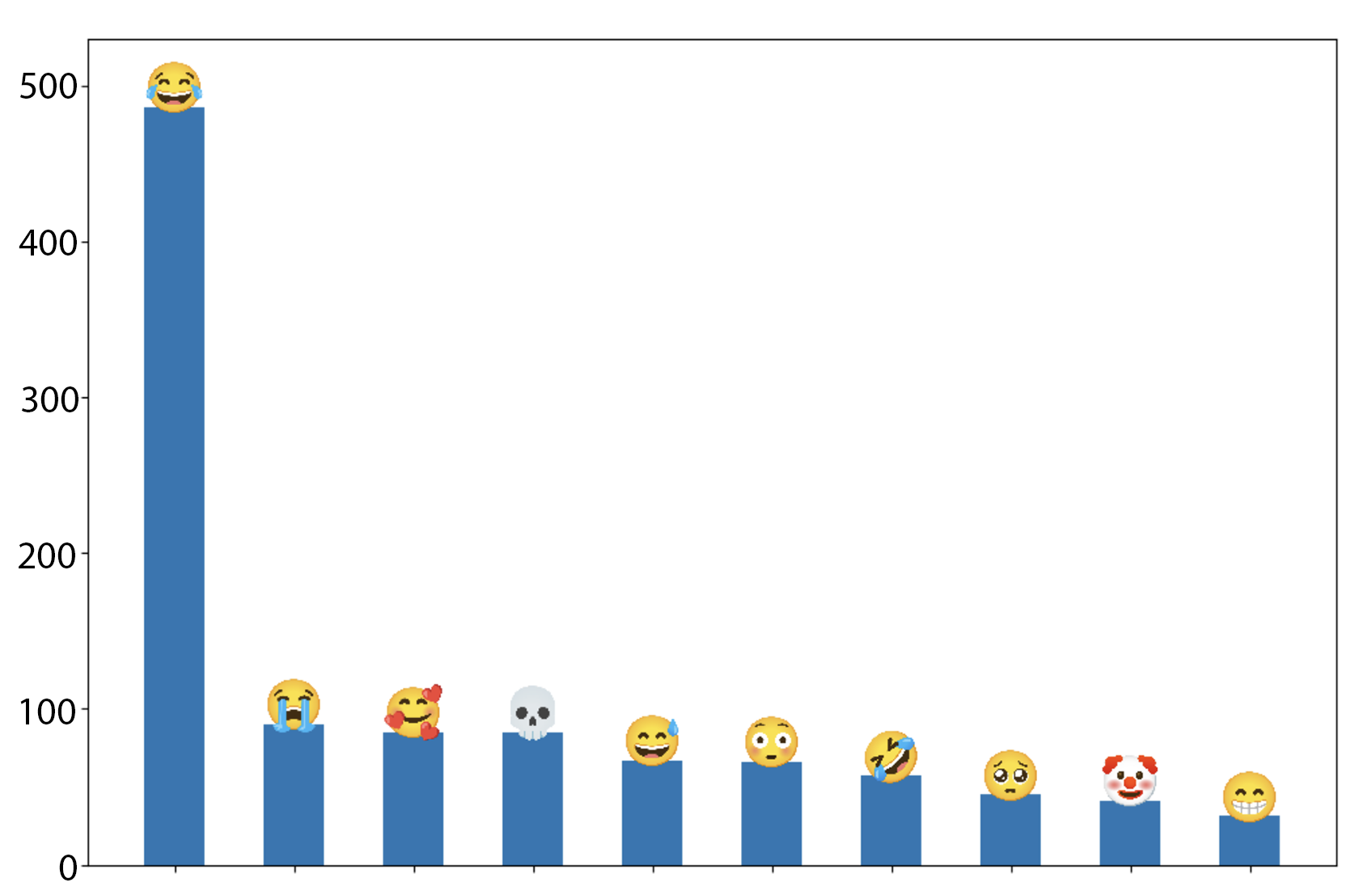}
		 \caption{offensive and non-offensive comments}
         \label{fig:most-freq-used-emojis-barchart}
     \end{subfigure}
     \begin{subfigure}[b]{0.48\textwidth}
         \centering
        \includegraphics[scale=0.24]{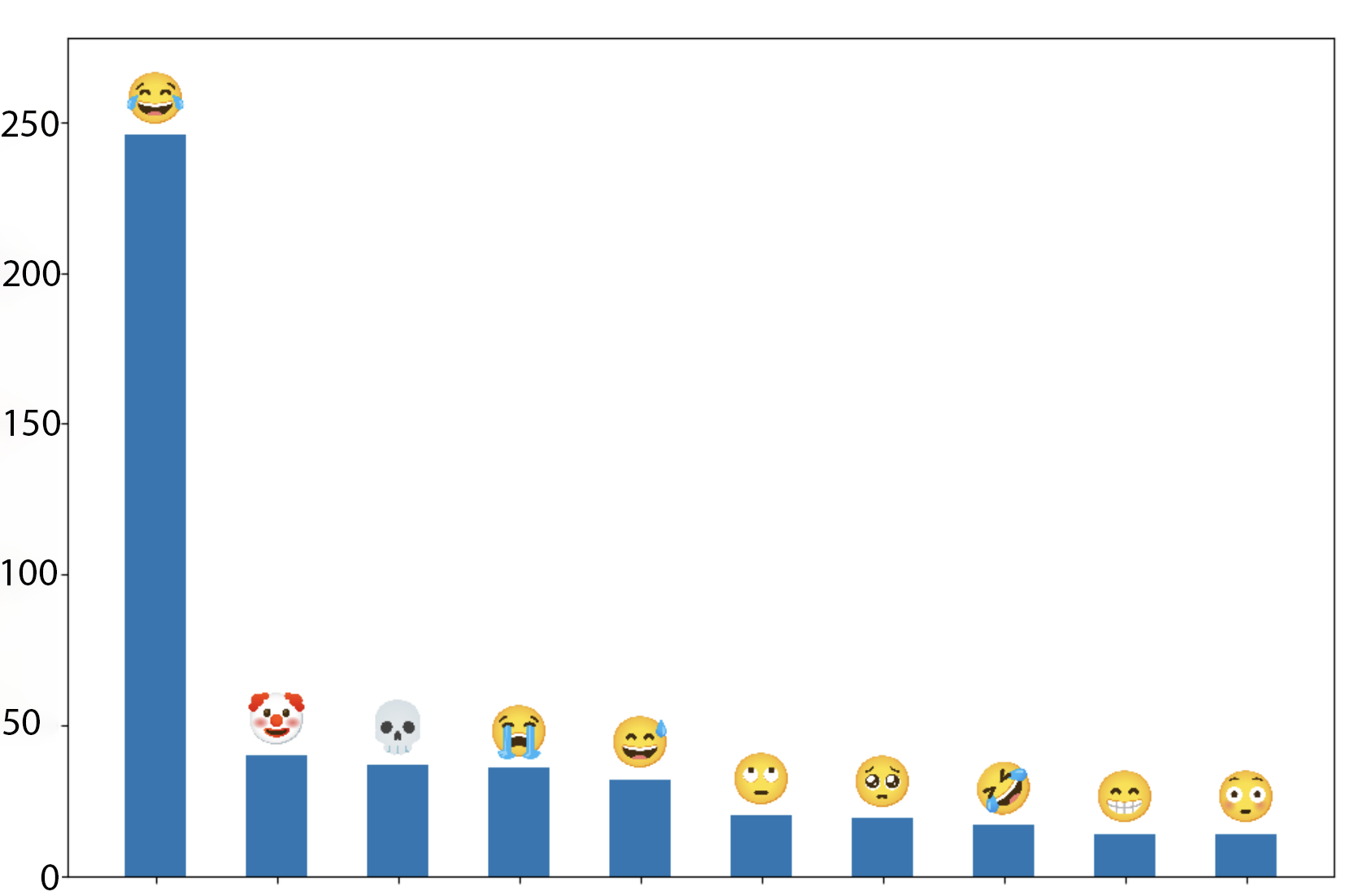}
		\caption{offensive comments}
		\label{fig:most-freq-used-offensive-emojis-barchart}
     \end{subfigure}
        \caption{Most frequently used emojis}
        \label{fig:most-frequently-used-emojis}
\end{figure*}

\subsection{Length distribution}\label{section:document-features}

Another metric to take into account for the dataset is the comment length. When observing Figures~\ref{fig:general-comment-length-distribution} and~\ref{fig:offensive-comment-length-distribution}, which depict the overall character comment length and the length of offensive comments in the corpus respectively, comments appear to be either rather short or long. 
Figure~\ref{fig:general-comment-length-distribution} shows that most of the comments have a length smaller or equal to $60$ characters. Comments shorter than $10$ characters usually show a user reacting to another comment with a single smiley or a single word, as shown by these examples: ``LOL'', ``brutal'', ``<33'', ``cap'', ``poser'', ``dumb''

In comparison, the comment length distribution for only the offensive documents, depicted in Figure~\ref{fig:offensive-comment-length-distribution}, shows that the length distribution is more evenly spread. Shorter messages are more common than longer ones, but the slump in the graph is not as significant as in the previous chart (Figure~\ref{fig:general-comment-length-distribution}).
In general, the data shows that offensive comments are more likely to include more text.


\subsection{The use of emojis}
For users, a modality of expression and creativity is the use of emojis~\cite{seemiller2018generation}. Within the TikTok domain, further examination of emoji relevance and use in offensive language is therefore important.
To this end, the frequency of occurrence of a certain (textual) emoji in the corpus was analyzed and plotted, resulting in the emoji cloud depicted in Figure~\ref{fig:most-freq-used-emojis-cloud}. Here it is important to note that both emoticons (textual emojis) and emojis are used to express emotions and were encountered during data collection. On this behalf, all emoticons were converted to their textual emoji counterpart before aggregating the results and creating the final charts in Figure~\ref{fig:most-freq-used-emojis-cloud} and~\ref{fig:most-freq-used-offensive-emojis-cloud}.


These clouds show that certain emojis such as the ``face with tears of joy''and ``loudly crying face'' are common and frequent to all examples, both offensive and non-offensive. 
%
In contrast, there is a clear increase in the use of certain emojis, such as the ``clown'' emoji and  ``skull''  emoji, in the offensive examples. This research shows that the  ``clown'' emoji  is often used to highlight foolish behavior: \\
		\includegraphics[scale=0.13]{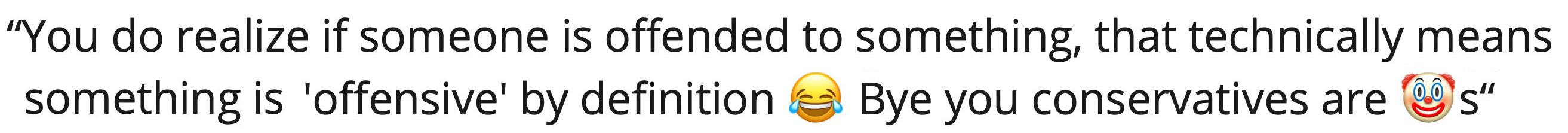}
The ``face with rolling eyes'' emoji, commonly used when expressing disapproval, is another seemingly more important emoji in offensive language: \\
		\includegraphics[scale=0.11]{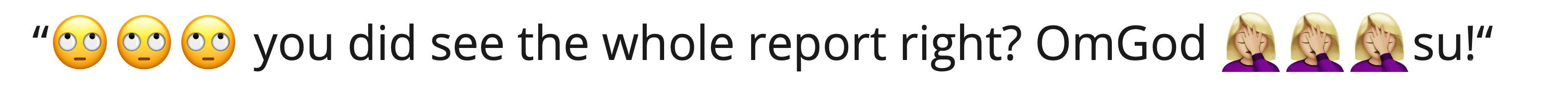} 

%
For this specific dataset, \(8.80\%\) of comments were found to contain one or more emojis. Subsequently, only 7.18\% of all offensive comments and 10.42\% of non-offensive comments contained emojis.



\section{Model development and design}\label{section:design}

This research involved creating binary classifiers using BERT, LR, and NB. BERT, a model for pre-training bidirectional representations from unlabeled data, was chosen in its ‘bert-base-uncased’ version considering the frequent disregard for punctuation and capital letters in online communication~\cite{hasyim2019linguistic}.

The compiled dataset served as a basis for examining the inherent linguistic characteristics of offensive language commonly encountered on the platform. The BERT model was then adapted to include tokens reflective of this specific context, like TikTok slang (‘Simp’, ‘Boomer’, ‘cap’)~\cite{nayak-etal-2020-domain}, and emojis, tokenized with the \textbf{emoji} and \textbf{emot} libraries~\cite{delobelle2019time}. This reduced subword tokenizations and enhanced the model’s performance.
The maximum sequence length was set to $150$, mirroring TikTok’s maximum comment length. We experimented with varying epochs, batch sizes, and learning rates~\cite{devlin2018bert}, and tested different training test/validation set ratios~\cite{browne2000cross}. The $80:20$ ratio yielded better model performance, so we used a validation ratio of $0.2$, split equally for testing and validation. Subsequently, multiple training cycles were conducted of which the best performing model was used as a reference. This iterative process helped in fine-tuning the model and achieving the best performance possible for this dataset. The training, test and validation subsets were selected randomly in each training cycle to ensure that we did not always use the same datasets during training. The experiments involved several preprocessing steps during training: (1) stopword removal, (2) emoji encoding, (3) lowercasing, (4) lemmatization, (5) punctuation removal. Lowercasing and emoji encoding were performed to match the preprocessing steps used to obtain the best possible fine-tuned BERT model. Previous research shows that the encoding of emojis for ML algorithms, such as LR and NB, can be done in a similar way as for the BERT model. The main difference in this approach is the way the ML algorithms will calculate the importance of particular words in the corpus: contrary to the emoji encoding for BERT, the emojis are replaced by their textual counterparts without additional characters~\cite{leCompteTravis2017}. As with BERT, documents in ML algorithms need to be tokenized in order for them to better interpret the meaning or relevance of a certain word in a particular document. To this extent, this research used TF-IDF for weighting the importance of tokens in the ML algorithms~\cite{ikonomakis2005text}.


\section{\label{section:results}Results}
\begin{table*}
    \footnotesize
    \centering
    \caption{Overview of model scores for the different variations}
    \begin{tabular}{|c|rcc|c|c|c|c|c|}
    \hline
    \multicolumn{1}{|l|}{Model variation}        & \multicolumn{3}{l|}{Confusion matrix}                                                  & \multicolumn{1}{l|}{F1}          & \multicolumn{1}{l|}{Accuracy} & \multicolumn{1}{l|}{Precision} & \multicolumn{1}{l|}{Recall} & \multicolumn{1}{l|}{Specificity} \\ \hline
    \multirow{3}{*}{Naive Bayes Default}         & \multicolumn{1}{l}{}               & Offensive                         & Not offensive & \multirow{3}{*}{\textbf{0.7063}} & \multirow{3}{*}{0.6547}       & \multirow{3}{*}{0.8733}        & \multirow{3}{*}{0.5929}     & \multirow{3}{*}{0.7991}          \\ \cline{3-4}
                                                & \multicolumn{1}{r|}{Offensive}     & \multicolumn{1}{c|}{\textbf{337}} & \textbf{70}   &                                  &                               &                                &                             &                                  \\ \cline{3-4}
                                                & \multicolumn{1}{r|}{Not Offensive} & \multicolumn{1}{c|}{\textbf{55}}  & \textbf{352}  &                                  &                               &                                &                             &                                  \\ \hline
    \multirow{3}{*}{Naive Bayes Emojis}          & \multicolumn{1}{l}{}               & Offensive                         & Not offensive & \multirow{3}{*}{\textbf{0.7177}} & \multirow{3}{*}{0.6744}       & \multirow{3}{*}{0.8708}        & \multirow{3}{*}{0.6105}     & \multirow{3}{*}{0.8091}          \\ \cline{3-4}
                                                & \multicolumn{1}{r|}{Offensive}     & \multicolumn{1}{c|}{\textbf{337}} & \textbf{70}   &                                  &                               &                                &                             &                                  \\ \cline{3-4}
                                                & \multicolumn{1}{r|}{Not Offensive} & \multicolumn{1}{c|}{\textbf{55}}  & \textbf{352}  &                                  &                               &                                &                             &                                  \\ \hline
    \multirow{3}{*}{Logistic Regression Default} & \multicolumn{1}{l}{}               & Offensive                         & Not offensive & \multirow{3}{*}{\textbf{0.7078}} & \multirow{3}{*}{0.7149}       & \multirow{3}{*}{0.7260}        & \multirow{3}{*}{0.6904}     & \multirow{3}{*}{0.7395}          \\ \cline{3-4}
                                                & \multicolumn{1}{r|}{Offensive}     & \multicolumn{1}{c|}{\textbf{331}} & \textbf{100}  &                                  &                               &                                &                             &                                  \\ \cline{3-4}
                                                & \multicolumn{1}{r|}{Not Offensive} & \multicolumn{1}{c|}{\textbf{104}} & \textbf{280}  &                                  &                               &                                &                             &                                  \\ \hline
    \multirow{3}{*}{Logistic Regression Emojis}  & \multicolumn{1}{l}{}               & Offensive                         & Not offensive & \multirow{3}{*}{\textbf{0.7095}} & \multirow{3}{*}{0.7174}       & \multirow{3}{*}{0.7260}        & \multirow{3}{*}{0.6938}     & \multirow{3}{*}{0.7408}          \\ \cline{3-4}
                                                & \multicolumn{1}{r|}{Offensive}     & \multicolumn{1}{c|}{\textbf{332}} & \textbf{99}   &                                  &                               &                                &                             &                                  \\ \cline{3-4}
                                                & \multicolumn{1}{r|}{Not Offensive} & \multicolumn{1}{c|}{\textbf{102}} & \textbf{282}  &                                  &                               &                                &                             &                                  \\ \hline
    \multirow{3}{*}{BERT Default}                & \multicolumn{1}{l}{}               & Offensive                         & Not offensive & \multirow{3}{*}{\textbf{0.8509}} & \multirow{3}{*}{0.8483}       & \multirow{3}{*}{0.8394}        & \multirow{3}{*}{0.8628}     & \multirow{3}{*}{0.8341}          \\ \cline{3-4}
                                                & \multicolumn{1}{r|}{Offensive}     & \multicolumn{1}{c|}{\textbf{337}} & \textbf{70}   &                                  &                               &                                &                             &                                  \\ \cline{3-4}
                                                & \multicolumn{1}{r|}{Not Offensive} & \multicolumn{1}{c|}{\textbf{55}}  & \textbf{352}  &                                  &                               &                                &                             &                                  \\ \hline
    \multirow{3}{*}{BERT Emojis}                 & \multicolumn{1}{l}{}               & Offensive                         & Not offensive & \multirow{3}{*}{\textbf{0.8564}} & \multirow{3}{*}{0.8469}       & \multirow{3}{*}{0.8137}        & \multirow{3}{*}{0.9039}     & \multirow{3}{*}{0.8119}          \\ \cline{3-4}
                                                & \multicolumn{1}{r|}{Offensive}     & \multicolumn{1}{c|}{\textbf{322}} & \textbf{85}   &                                  &                               &                                &                             &                                  \\ \cline{3-4}
                                                & \multicolumn{1}{r|}{Not Offensive} & \multicolumn{1}{c|}{\textbf{40}}  & \textbf{367}  &                                  &                               &                                &                             &                                  \\ \hline
    \multirow{3}{*}{BERT Slang}                  & \multicolumn{1}{l}{}               & Offensive                         & Not offensive & \multirow{3}{*}{\textbf{0.8573}} & \multirow{3}{*}{0.8551}       & \multirow{3}{*}{0.8487}        & \multirow{3}{*}{0.8662}     & \multirow{3}{*}{0.8461}          \\ \cline{3-4}
                                                & \multicolumn{1}{r|}{Offensive}     & \multicolumn{1}{c|}{\textbf{318}} & \textbf{89}   &                                  &                               &                                &                             &                                  \\ \cline{3-4}
                                                & \multicolumn{1}{r|}{Not Offensive} & \multicolumn{1}{c|}{\textbf{55}}  & \textbf{352}  &                                  &                               &                                &                             &                                  \\ \hline
    \multirow{3}{*}{BERT Emoji \& slang}         & \multicolumn{1}{l}{}               & Offensive                         & Not offensive & \multirow{3}{*}{\textbf{0.8633}} & \multirow{3}{*}{0.8518}       & \multirow{3}{*}{0.8146}        & \multirow{3}{*}{0.9182}     & \multirow{3}{*}{0.8077}          \\ \cline{3-4}
                                                & \multicolumn{1}{r|}{Offensive}     & \multicolumn{1}{c|}{\textbf{232}} & \textbf{199}  &                                  &                               &                                &                             &                                  \\ \cline{3-4}
                                                & \multicolumn{1}{r|}{Not Offensive} & \multicolumn{1}{c|}{\textbf{53}}  & \textbf{331}  &                                  &                               &                                &                             &                                  \\ \hline
    \end{tabular}
    \label{table:model-variations}
\end{table*}
This section presents a comprehensive analysis of the results obtained from the iterative steps outlined in the preceding sections. The ensuing discussion will focus on the performance of the BERT model and its fine-tuned variants, in comparison to the baseline BERT model without any pre-processing steps. The latter part of this section explores the outcomes of the training sessions involving the LR and NB algorithms. The outcomes from the various fine-tuning stages are discussed in detail. Furthermore, the performance metrics such as F1 score, accuracy, precision, recall, and specificity for different model variations are presented in Table~\ref{table:model-variations}. One of the key highlights is the F score of the default BERT model training, which stands at $0.8509$, even in the absence of any pre-processing. These scores are derived from the values in the confusion matrices and indicate that the application of both emoji and slang tokenization result in a marginal enhancement of the model’s performance, with increments of $0.0055$ and $0.0064$, respectively. Moreover, the model that delivers the best performance is the one that incorporates both emoji and slang tokenization. This strategy leads to an overall performance improvement of $0.0124$ when compared to the baseline model.

Nevertheless, it is important to note that, as per the analysis, only $8.80$\% of comments in the corpus contained emojis. Despite the balance between offensive and non-offensive text, the relatively scarce presence of emojis across text accounts for the modest improvement observed after implementing emoji tokenization. Another metric that underscores this conclusion is the recall measurement, which shows the model’s efficacy in identifying offensive content. As indicated in Table~\ref{table:model-variations}, the combination of emoji and slang tokenization yields the most favorable results with a recall score of ($0.9182$). Hence, the results reveal that the tokenization of emojis ($0.9039$) have a more substantial influence on the recall score than slang tokenization ($0.8662$). 
Furthermore, two variations of the baseline machine learning models were trained for comparative analysis. Table~\ref{table:model-variations} presents the F scores for both the default model, which employs TF-IDF tokenization and standard pre-processing, and a variant where emojis are encoded and tokenized instead of being removed. It can be observed that both models yield lower F scores than the fine-tuned BERT model earlier discussed. While the F scores are reasonably good, the recall score stands at $0.5929$ without emoji encoding and slightly higher at $0.6105$ when emojis are considered for the NB algorithm. Considering the total number of True offensive comments, as shown in Table~\ref{table:model-variations}, the performance of this ML algorithm is evidently inferior compared to the $0.9182$ recall score of the fine-tuned BERT model. These performance differences can be attributed to the contextual features that a comprehensive language model like BERT can capture, as opposed to ML algorithms such as NB. This is also evident in tasks like sarcasm detection, where understanding semantic relationships between words is crucial~\cite{mukherjee2017sarcasm}. The pre-processing steps and model variations applied to the LR algorithm mirror those used for NB. As shown in Table~\ref{table:model-variations}, the F scores for the default model using TF-IDF tokenization and the model incorporating emoji tokenization are presented. When compared with NB results in Table~\ref{table:model-variations}, it can be seen that the F score is better for NB when compared to LR, with scores of $0.7177$ and $0.7095$ respectively. The model’s performance sees an uptick between the default variant and the one that includes emoji encoding, a trend that aligns with observations from the other experiments. In terms of recall, both versions of the LR algorithm demonstrate comparable efficacy in accurately detecting offensive language. In conclusion, the fine-tuned BERT model surpasses the performance of baseline ML algorithms such as NB and LR. The results further underscore that fine-tuning enhances the overall performance of the model.

\section{Discussion and Conclusions\label{section:conclusion}}

The proliferation of offensive content in various social media platforms has become a significant societal concern as its impact transcends the digital borders. Beyond the immediate implications on users’ experience and platforms’ integrity, the increase and normalization of offensive behavior and language can imply far-reaching consequences like the creation and perpetuation of harmful stereotypes, discriminatory attitudes and racism, social division, and erosion of well-established socio-ethical principles, norms, and values. For TikTok, this represents a critical issue seeing the average young age of its users plus a series of technological, social, and psychological factors that play a role in this process. To name a few, the ease of content creation and dissemination, anonymity of users, viral nature of the algorithms used, and the echo chamber effect that implies exposing users to content aligned with their existing preferences learned from their views. Hence, addressing the proliferation of offensive content is not merely a matter of platform governance, but implies a broader societal effort from relevant policy makers, technological companies, practitioners, and researchers to promote and foster digital literacy, and avoid or at least mitigate the negative effects. 

Releasing more publicly open datasets, and building effective AI models play a crucial role in building efforts for content moderation and countering the proliferation of offensive content on social media platforms like TikTok. To this end, this research proposes a TikTok dataset and a series of deep learning and machine learning models for detecting offensive content on TikTok. From the analysis conducted, it was found that a combination of emojis (“skull” and “crying your heart out”) and custom vocabulary (e.g., “simp”, “cap”, “boomer”) are sometimes used to express emotions and convey offensive content. Furthermore, a series of deep learning and machine learning models are built to detect offensive language using three variants of BERT (i.e., with emoji tokenization, with slang tokenization, with both slang and emoji tokenization), logistic regression, and na{\"\i}ve Bayes fine-tuned algorithms. From the results obtained, the BERT model with emoji and slang tokenization was the best-performing one, revealing the fact that tokenization of slang, emojis and emoticons can improve the model’s ability to predict offensive content correctly. Building upon these promising outcomes, the following further perspectives are seen. Firstly, exploring the integration of multimodal features like image and video analysis alongside textual data to gather a more holistic understanding of offensive content dynamics. Secondly, building collaborative efforts between field experts from various domains allows the adoption of a transdisciplinary effort facilitating the development of AI-based and gamification modeling and simulation solutions for awareness, educational, and decision-making support purposes. 


\bibliographystyle{IEEEtran}
\bibliography{bibliography}

\end{document}